\documentclass[runningheads]{llncs}
\usepackage[T1]{fontenc}
\usepackage{graphicx}
\usepackage{makecell}
\usepackage{wrapfig}
\usepackage{epsfig}
\usepackage{algorithm}
\usepackage{algorithmic}
\usepackage{amsmath}
\usepackage{amssymb}
\usepackage{mathtools}
\usepackage{multirow}
\usepackage{subfigure}
\usepackage[capitalize,noabbrev]{cleveref}
\usepackage[misc]{ifsym}
\usepackage{amsmath,amsfonts,bm}

\def\eqref#1{equation~\ref{#1}}

\def\1{\bm{1}}

\def\vtheta{{\bm{\theta}}}

\def\vr{{\bm{r}}}
\def\vs{{\bm{s}}}

\def\vw{{\bm{w}}}
\def\vx{{\bm{x}}}
\def\vy{{\bm{y}}}
\def\vz{{\bm{z}}}

\def\mA{{\bm{A}}}

\def\mT{{\bm{T}}}

\DeclareMathAlphabet{\mathsfit}{\encodingdefault}{\sfdefault}{m}{sl}
\SetMathAlphabet{\mathsfit}{bold}{\encodingdefault}{\sfdefault}{bx}{n}

\def\gG{{\mathcal{G}}}

\def\gR{{\mathcal{R}}}

\def\gX{{\mathcal{X}}}

\newcommand{\R}{\mathbb{R}}

\title{Score-based Generative Models for Photoacoustic Image Reconstruction with Rotation Consistency Constraints}
\titlerunning{PAT Reconstruction with RCC-SGM}
\author{Shangqing Tong\inst{1}\thanks{The first two authors have contributed equally to this paper.} \and
Hengrong Lan\inst{2}$^{\star}$ \and
Liming Nie\inst{3} \and
Jianwen Luo\inst{2}$^{(\textrm{\Letter})}$ \and
Fei Gao\inst{1}$^{(\textrm{\Letter})}$}
\authorrunning{S. Tong \& H. Lan et al.}
\institute{ShanghaiTech University, Shanghai 201210, China \\
\email{gaofei@shanghaitech.edu.cn} \and
Tsinghua University, Beijing 100084, China \\
\email{luo\_jianwen@tsinghua.edu.cn} \and
Guangdong Academy of Medical Sciences, Guangzhou 510000, China}
\begin{document}
\maketitle              
\begin{abstract}
Photoacoustic tomography (PAT) is a newly emerged imaging modality which enables both high optical contrast and acoustic depth of penetration. Reconstructing images of photoacoustic tomography from limited amount of senser data is among one of the major challenges in photoacoustic imaging.
Previous works based on deep learning were trained in supervised fashion, which directly map the input partially known sensor data to the ground truth reconstructed from full field of view.
Recently, score-based generative models played an increasingly significant role in generative modeling.
Leveraging this probabilistic model, we proposed \emph{Rotation Consistency Constrained Score-based Generative Model (RCC-SGM)}, which recovers the PAT images by iterative sampling between Langevin dynamics and a constraint term utilizing the rotation consistency between the images and the measurements.
Our proposed method can generalize to different measurement processes (32.29 PSNR with 16 measurements under random sampling, whereas 28.50 for supervised counterpart), while supervised methods need to train on specific inverse mappings.
\end{abstract}
\section{Introduction}\label{sec:intro}

Photoacoustic tomography (PAT)~\cite{Xu2006PhotoacousticII}, which enables high optical absorption contrast and penetration depth, is a newly emerged imaging modality based on photoacoustic effect.
When biological tissues are exposed to non-ionizing laser radiation, the tissues absorb the energy of the laser and convert the energy to heat, which leads to local transient thermoelastic expansion.
During this procedure, wide band ultrasound signals are emitted, carrying knowledge about the structural and functional information of the tissues. The ultrasound signals, so called photoacoustic signals, are then captured with ultrasound transducers, and will be further analysed to produce images with reconstruction algorithms.

Traditional reconstruction algorithms of PAT include back projection (BP) algorithm~\cite{Xu2005UniversalBA} and delay-and-sum (DAS) algorithm.
However, these approaches are vulnerable to the detection angles and numbers of measurements. Besides, several model-based iterative methods, such as total variation~\cite{Chambolle2004AnAF} and wavelet~\cite{Chang2000AdaptiveWT}, are also used for undersampled PAT image enhancement.

With the in-depth development of deep learning theory, it has been widely used to perform undersampled reconstructions.
To our best knowledge, previous studies with deep learning methods are mostly in an end-to-end supervised fashion, which directly trained a well-designed deep neural network with a pre-generated paired dataset.
Ref.~\cite{Davoudi2019DeepLO,Lan2019KiGANKI,Guan2019LimitedViewAS,Lan2020YNetHD} utilized deep learning methods to reconstruct images, while Ref.~\cite{Hauptmann2017ModelBasedLF} combined deep learning with model-based approaches. Ref.~\cite{Meng2022ANU} also utilized diffusion model for multi-modal medical image completion.
However, paired datasets either are expensive to obtain, or do not exist under some particular circumstances. Though supervised solutions provide high quality images, it hinders the generalization capacity, leading to some counterintuitive problems such as worse reconstruction results with even more measurements.

\begin{figure}[t]
    \centering
    \includegraphics[width=\textwidth,trim=0 52 0 52,clip]{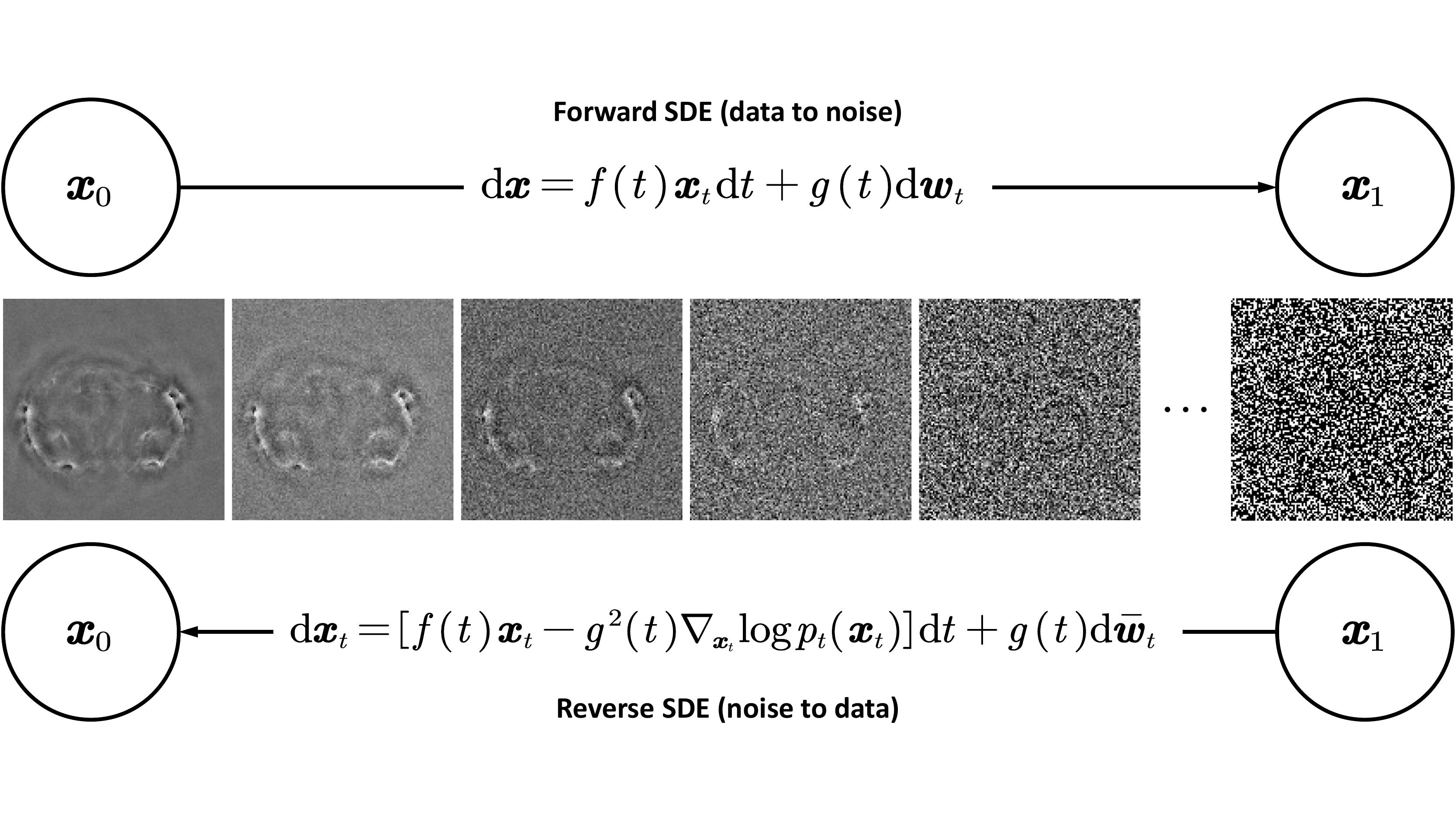}
    \caption{Forward and reverse process of our proposed method. We first perturb the data images into a prior distribution by the trajectory of an stochastic differential equation (SDE). We generate samples via solving the reverse SDE by some sampling algorithms.}
\end{figure}

With the emergence of score-based generative models (SGM)~\cite{Song2019GenerativeMB,Song2020ScoreBasedGM}, a wide range of methods based on SGM have been proposed to solve inverse problems in medical imaging in recent years.
Ref.~\cite{Jalal2021RobustCS} first proposed a Langevin solver to produce MRI images by adding gradients of likelihood. After that, several works had been proposed to improve the performance of diffusion models for inverse problems in medical imaging~\cite{Chung2021ScorebasedDM,Chung2022ImprovingDM,Chung2021ComeCloserDiffuseFasterAC}. Yang Song et al.~proposed the first sampler to solve inverse problems in CT~\cite{Song2021SolvingIP}.
These methods are mostly proposed to solve inverse problems in MRI and CT. To the best of our knowledge, there are no existing unsupervised methods for PAT reconstruction.

In this work, we propose the \emph{Rotation Consistency Constrained Score-based Generative Model (RCC-SGM)} for PAT reconstruction. Our work, which utilized the SGM to solve inverse problem, is \emph{the first} fully unsupervised method that has witnessed success on \emph{in vivo} PAT data.
We have shown that by imposing the rotation consistency between the measurements and images, an unsupervised SGM can achieve competitive results of reconstructing photoacoustic images to supervised ones.
Specifically, using a trained generative prior, we proposed a solver to inject measurement information into the process of solving the reverse stochastic differential equation (SDE). Experiment results are evaluated both quantitatively and qualitatively.
We achieved higher PSNR than the supervised UNet baselines. The evaluation showed that with random sampling of 16 measurements, we obtained 32.29 PSNR, while UNet obtained 28.50 PSNR; with limited view of 64 measurements, our method obtained 37.21 PSNR and 0.975 SSIM, while UNet obtained 31.31 PSNR and 0.970 SSIM.

\section{Background}\label{sec:bg}

\subsection{Photoacoustic Tomography}

Photoacoustic imaging is an imaging modality based on photoacoustic effect~\cite{Xu2006PhotoacousticII}. Given a heat source $H(\vr, t)$, the pressure $p(\vr, t)$ in an acoustically homogeneous inviscid medium is given by
\begin{equation}
    \nabla^2 p(\vr, t) - \frac{1}{v_s^2} \frac{\partial^2}{\partial t^2} p(\vr, t) = - \frac{\beta}{C_p} \frac{\partial}{\partial t} H(\vr, t), \label{eq:pa-pressure}
\end{equation}
where $\beta$ is the isobaric volumn expansion coefficient, $v_s$ is the speed of sound in the medium, and $C_p$ is the specific heat. The forward solution of~\cref{eq:pa-pressure} in time domain can be achieved via Green's function,
\begin{equation}
    p(\vr, t) = \frac{\beta}{4\pi C_p} \iiint \frac{\mathrm{d} \vr'}{\left| \vr - \vr' \right|} \frac{\partial H(\vr', t')}{\partial t'} \bigg|_{t' = t - \left| \vr - \vr' \right| / v_s} .
\end{equation}
The magnitude of the received photoacoustic signal is proportional to the local energy deposition, which illustrates the physiological optical absorption contrast.

\subsection{Score-based Generative Modeling}

Yang Song, et al.~proposed SGMs~\cite{Song2019GenerativeMB,Song2020ScoreBasedGM}, which produce samples via Langevin dynamics using gradients of the logarithmic data distribution estimated by score matching.
In general, an SGM perturbs the training data with Gaussian noise during the forward (perturbation) process following the forward stochastic differential equation (SDE),
\begin{equation}
    \mathrm{d} \vx_t = f (t) \vx_t \mathrm{d} t + g(t) \mathrm{d} \vw_t, \label{eq:forward-sde}
\end{equation}
where $\vw_t$ is a standard Wiener process, $t \in \left[0, 1\right]$, $f (t): \left[0, 1\right] \to \R$ is called the drift coefficient of $\vx$ and $g(t): \left[0, 1\right] \to \R$ is called the diffusion coefficient.
A noise conditional score network is trained to capture the gradient of the logarithmic data distribution (\textit{i.e.} score function, $\nabla_{\vx} \log p (\vx)$) solving the denoising score matching objective function~\cite{Vincent2011ACB}. In particular, the Variance Exploding (VE) SDE~\cite{Song2020ScoreBasedGM} is defined as
\begin{equation}
    \mathrm{d} \vx_t = \sqrt{\frac{\mathrm{d}\left[ \sigma^2 (t) \right]}{\mathrm{d} t}} \mathrm{d} \vw_t,
\end{equation}
which provides a stochastic process with exploding variance with $t\to +\infty$. Other SDEs include the Variance Preserving (VP) SDE~\cite{Song2020ScoreBasedGM} and sub-VP SDE~\cite{Song2020ScoreBasedGM}.

The reverse process traces back from drawing a sample from the prior distribution and recover the deserved samples step by step via gradually removing noise using the estimated score.
Theoretically, the corresponding reverse process of~\cref{eq:forward-sde} is
\begin{equation}
    \mathrm{d} \vx_t = \left[ f (t) \vx_t - g^2(t) \nabla_{\vx_t} \log p_t (\vx_t) \right] \mathrm{d} t + g(t) \mathrm{d} \bar{\vw}_t, \label{eq:reverse-sde}
\end{equation}
where $\bar{\vw}_t$ is a standard Wiener process in reverse direction, and $\mathrm{d} t$ is an infinitesimal negative time step, $t \in \left[0, 1\right]$.
The term $\nabla_{\vx_t} \log p_t (\vx_t)$ is called the score function of $p_t (\vx _t)$. By solving the reverse time SDE, we can obtain samples with only the score of a distribution.
There are various methods to solve SDEs. 
Yang Song, et al.~proposed annealed Langevin dynamics (ALD)~\cite{Song2019GenerativeMB} to approximate samples from a distribution by gradually removing noise from a noisy image extracted from a prior distribution (\emph{i.e.} normal distribution) with several times of noisy gradient ascent in a decreasing order of noise.
\section{Methodology}\label{sec:method}

\subsection{Problem Overview}

The basic inverse model of PAT imaging is the DAS algorithm. We used the model matrix, which is calculated following the curve-driven-based method~\cite{Liu2016CurveDrivenBasedAI}, denoted by $\mA \in \R ^{m\times n}$ and $\mA^\dagger \in \R ^{n\times m}$ respectively, to approximate the forward and inverse operators.
An image, denoted by $\vx \in \R^n$, is passed through the forward process. Measurements $\vy \in \R^m$ are obtained, but only a subset of the measurements which consists of $\tilde{n} < m$ known photoacoustic signals are accessible for reconstruction. The measurement process can be modeled through a linear equation,
\begin{equation}
    \vy = \mA \vx + \bm{\eta} \label{eq:inv-problem}
\end{equation}
where $\bm{\eta}$ is the noise in the measurement process. 

\subsection{Controllable Generation with Score-based Generative Models}

In this work, a noise conditional score network (NCSN) is trained to capture the scores of the distribution perturbed with $L$ different noise scales ranging from $\sigma_{\text{min}}$ to $\sigma_{\text{max}}$, denoted by $\vs_{\vtheta} (\vx, \sigma_i)$ for $i=1, 2, \ldots, L$. During the annealed Langevin dynamics (ALD) algorithm, the gradient ascent step is performed $T$ times for a single noise scale. The outputs of ALD under various noise scales, denoted by $\left\{ \vx_i \right\} _{i=1}^L$, can be seen as an estimation of the sample perturbed with a sequence of noise. The last sample, $\vx_0$, produced with the ALD under noise scale $\sigma_{\text{min}}$ can be treated as an certain approximation of reconstruction.

However, the reconstruction from limited number of measurements can be treated as sampling from a conditioned distribution $p(\vx \mid \vy)$. To sidestep this intractable item, we can refer to the Bayes' rule, \textit{i.e.}
\begin{equation}
    \nabla_{\vx} \log p(\vx \mid \vy) = \nabla_{\vx} \log p(\vx) + \nabla_{\vx} \log p(\vy \mid \vx),\label{eq:bayes}
\end{equation}
where $\nabla_{\vx} \log p(\vx)$ can be replaced by the NCSN we trained, and the gradient of likelihood $\nabla_{\vx} \log p(\vy \mid \vx)$ is relatively easy to obtain~\cite{Jalal2021RobustCS}. We iterate the ALD from $\sigma_{\text{max}}$ to $\sigma_{\text{min}}$ by updating
\begin{equation}
    \vx^{(t+1)}_{i} \leftarrow \vx^{(t)}_{i} + \epsilon \nabla_{\vx^{(t)}_{i}} \log p(\vx^{(t)}_{i} \mid \vy) + \sqrt{2\epsilon}\vz^{(t)},\label{eq:ALD}
\end{equation}
for $t=0, 1, \ldots, T-1$ at the $i$-th noise scale. By plugging in~\cref{eq:bayes}, we can inject the mapping between the regions to the sampling process, thus produce reconstructions.

\subsection{Rotation Consistency Constraints}

The main purpose of our idea is to modify the sampling algorithm of the SGM to gain better reconstruction results for PAT, which is achieved by regularizing the ALD with rotation consistency constraints between PAT measurements and images.

Given a signal set $\gX \subseteq \R^n$, a group of transformations $\{ \mT _g \} _{g=1} ^{|\gG|}$, notated by $\gG$, is equivariant to the signal set if $\forall \vx \in \gX$, $T_g \vx \in \gX$ for all transformations in the transformation group, \textit{i.e.}
\begin{equation}
    \mA \mT _g (\vx) = \mT _g (\mA \vx).
\end{equation}
Let $\gR$ be the group of rotations by the angle between two adjacent sensors ($360^{\circ} / m$ degree) where $\left| \gR \right| = m$. Considering the symmetry of measurements and images, $\mT _r$ is the rotation operator on images and $\tilde{\mT} _r$ is the shift operator for the sensor data, where $r \in \gR$.
For the full-sampled processes, the rotation transformation group should be equivariant to the model matrix, \textit{i.e.}
\begin{equation}
    \tilde{\mT_r} \left( \mA \vx \right) = \mA \mT_r \left( \vx \right). \label{eq:rot-eq}
\end{equation}
In ideal, full-sampled cases, there exists the consistency. However, the equivariance may not hold in each approximation $\vx_i$ the ALD has produced, thus this error may accumulate gradually as the iteration proceeds. To correct this deviation in every noise scale in ALD, we do
\begin{equation}
    \vx_{i}^{(T)} \leftarrow \vx_{i}^{(T)} + \alpha \mA^\dagger \left[ \tilde{\mT_r} \left( \mA \vx_{i}^{(T)} \right) - \mA \mT_r \left( \vx_{i}^{(T)} \right) \right]
\end{equation}
right after the $T$ steps of ALD update at the $i$-th noise scale described in~\cref{eq:ALD}, where $\alpha \in \left[ -1, 1 \right]$ is a hyperparameter to be tuned.
\section{Experiments}\label{sec:experiments}

\begin{figure}[t]
    \centering
    \includegraphics[width=\textwidth,trim=15 35 15 35,clip]{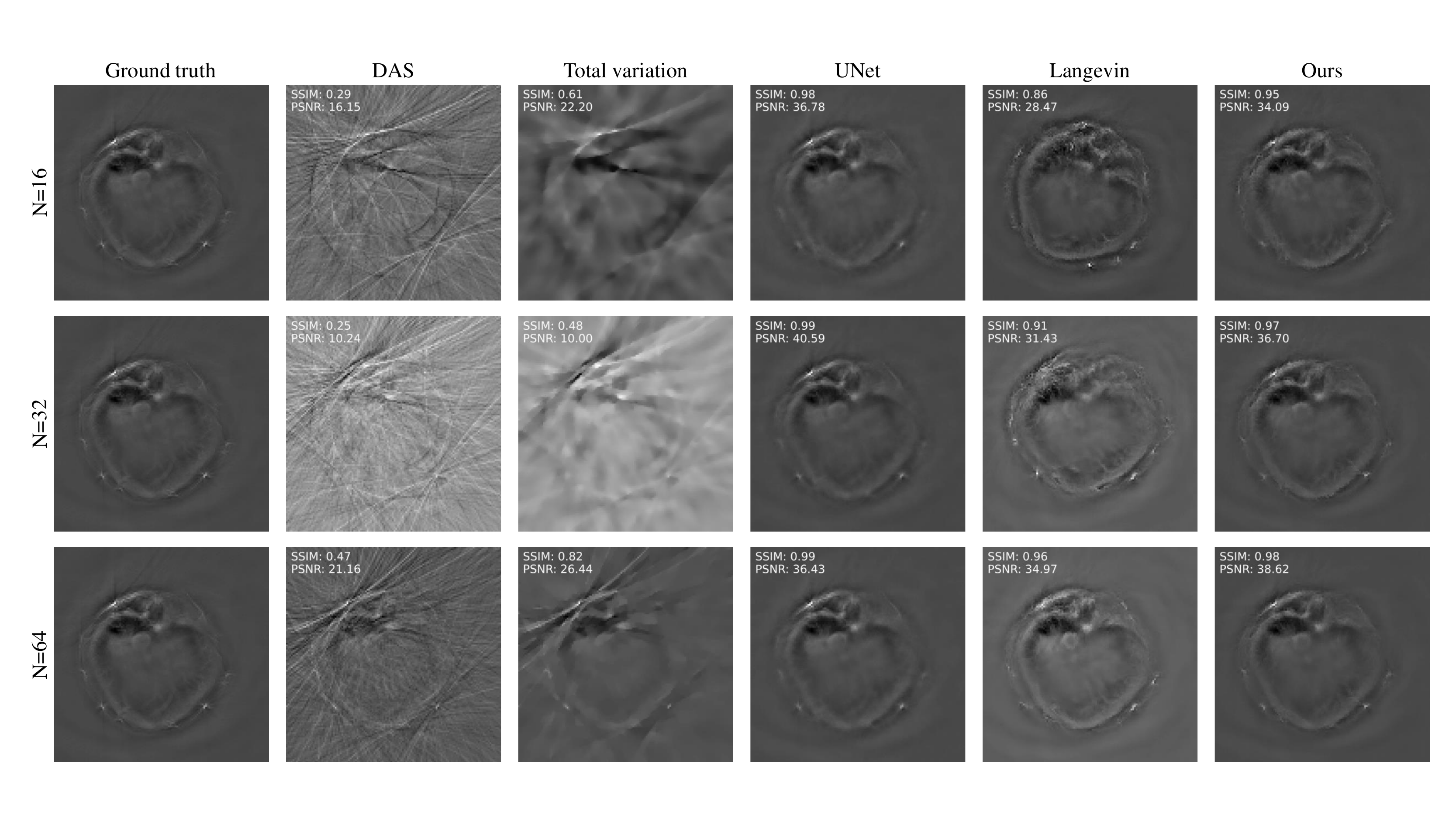}
    \caption{Experiment results of uniform sampling reconstruction in $128\times 128$.}\label{fig:results}
\end{figure}

\paragraph{Setup} Our experiments have assessed different sampling patterns, including uniform sampling, random sampling and limited view. Uniform sampling performs a uniform downsampling of PA signals. Random sampling stochastically chooses a deserved number of known measurements. Limited view chooses to leave a portion of the adjacent signals as known measurements while ignores the rest. In addition, reconstructions with different numbers of measurements ranging from 16 to 64 were also evaluated.

\paragraph{Dataset} The dataset we used is a collection of \emph{in vivo} PAT scans of mice. The scans were performed with a panoramic PAT system (SIP-PACT512, Uion Photoacoustic Technologies Co., Ltd., Wuhan, China). We used a 1064 nm laser to irradiate the mice, and the PA signals were then collected by a $360^{\circ}$ annular transducer array with 512 channels at 5 MHz center frequency. The sampling frequency was 40 MHz. Four healthy nude mice (8 weeks old, SPF Biotechnology Co., Ltd., Beijing, China) were involved in our experiment. Immersed in a homothermal water tank, these mice were scanned by moving the animal holder using a positioner to image the entire body in 0.02 mm steps. The experiments were approved by the Institutional Animal Care and Use Committee in Guangdong Provincial People's Hospital. The dataset contains 3382 scans, with 382 slices reserved for evaluation and the remaining 3000 samples used for training.

\paragraph{Standard techniques for PAT reconstruction} We included learning-free denoising methods for comparison. One of them is the linear reconstruction with no regularizations. The observations were recontructed using undersampled measurements by simply calculating $\mA^\dagger \vy$. The other is the total variation method. We chose $\lambda = 2000$, tolerance $\epsilon = 2\times 10^{-4}$ for the denoising. The iteration was performed up to 200 times.

\paragraph{Supervised learning baselines} UNet~\cite{Ronneberger2015UNetCN} 
is a simple baseline for undersampled PAT image denoising. 
The input undersampled signals are first converted to noisy images, and the UNets were trained to map the images to ground truths respectively for different measurement processes, optimized by AdamW optimizer~\cite{Loshchilov2017DecoupledWD}. The UNets were trained for 25000 iterations with batch size 16. The learning rate followed a cosine annealing schedule~\cite{Loshchilov2016SGDRSG} from $1\times 10^{-4}$ to $1\times 10^{-7}$.

\begin{table}[t]
    \centering
    \caption{Results for undersampled PAT reconstruction under different measurement processes}\label{tab:results}
    \begin{tabular}{l|c|cc|cc|cc}
    \Xhline{2\arrayrulewidth}
    \multicolumn{1}{c|}{\multirow{2}{*}{Methods}} & \multicolumn{1}{c|}{\multirow{2}{*}{Channels}} & \multicolumn{2}{c|}{Uniform Sampling}  & \multicolumn{2}{c|}{Random Sampling}   & \multicolumn{2}{c}{Limited View}  \\ \cline{3-8} 
    \multicolumn{1}{c|}{} & \multicolumn{1}{c|}{}  & \multicolumn{1}{c}{PSNR$\uparrow$} & \multicolumn{1}{c|}{SSIM$\uparrow$} & \multicolumn{1}{c}{PSNR$\uparrow$} & \multicolumn{1}{c|}{SSIM$\uparrow$} & \multicolumn{1}{c}{PSNR$\uparrow$} & \multicolumn{1}{c}{SSIM$\uparrow$} \\ \hline 
    Ours  & 16   & 32.05\scalebox{0.75}{$\pm$1.85}  & 0.924\scalebox{0.75}{$\pm$0.032}  & 32.29\scalebox{0.75}{$\pm$1.89}& 0.927\scalebox{0.75}{$\pm$0.032}    & 26.03\scalebox{0.75}{$\pm$1.88}& 0.810\scalebox{0.75}{$\pm$0.054}\\
    Langevin & 16   & 27.10\scalebox{0.75}{$\pm$1.43}  & 0.812\scalebox{0.75}{$\pm$0.048}  & 26.61\scalebox{0.75}{$\pm$1.46}& 0.800\scalebox{0.75}{$\pm$0.049}    & 24.33\scalebox{0.75}{$\pm$1.50}& 0.743\scalebox{0.75}{$\pm$0.053}\\
    UNet  & 16   & 32.85\scalebox{0.75}{$\pm$6.77}  & 0.970\scalebox{0.75}{$\pm$0.036}  & 28.50\scalebox{0.75}{$\pm$6.18}& 0.950\scalebox{0.75}{$\pm$0.043}    & 28.43\scalebox{0.75}{$\pm$6.27}& 0.950\scalebox{0.75}{$\pm$0.049}\\ \hline
    Ours  & 32   & 34.61\scalebox{0.75}{$\pm$1.68}  & 0.954\scalebox{0.75}{$\pm$0.017}  & 34.27\scalebox{0.75}{$\pm$1.60}& 0.953\scalebox{0.75}{$\pm$0.017}    & 29.31\scalebox{0.75}{$\pm$2.01}& 0.894\scalebox{0.75}{$\pm$0.039}\\
    Langevin & 32   & 30.52\scalebox{0.75}{$\pm$1.43}  & 0.900\scalebox{0.75}{$\pm$0.031}  & 30.28\scalebox{0.75}{$\pm$1.42}& 0.896\scalebox{0.75}{$\pm$0.032}    & 26.42\scalebox{0.75}{$\pm$1.85}& 0.821\scalebox{0.75}{$\pm$0.046}\\
    UNet  & 32   & 33.13\scalebox{0.75}{$\pm$7.05}  & 0.973\scalebox{0.75}{$\pm$0.032}  & 30.59\scalebox{0.75}{$\pm$6.54}& 0.964\scalebox{0.75}{$\pm$0.035}    & 30.31\scalebox{0.75}{$\pm$6.87}& 0.959\scalebox{0.75}{$\pm$0.043}\\ \hline
    Ours  & 64   & 36.79\scalebox{0.75}{$\pm$1.49}  & 0.971\scalebox{0.75}{$\pm$0.009}  & 36.60\scalebox{0.75}{$\pm$1.45}& 0.971\scalebox{0.75}{$\pm$0.009}    & 37.21\scalebox{0.75}{$\pm$1.30}& 0.975\scalebox{0.75}{$\pm$0.007}\\
    Langevin & 64   & 32.83\scalebox{0.75}{$\pm$1.46}  & 0.939\scalebox{0.75}{$\pm$0.019}  & 32.59\scalebox{0.75}{$\pm$1.52}& 0.936\scalebox{0.75}{$\pm$0.021}    & 32.87\scalebox{0.75}{$\pm$1.52}& 0.939\scalebox{0.75}{$\pm$0.019}\\
    UNet  & 64   & 33.02\scalebox{0.75}{$\pm$6.92}  & 0.975\scalebox{0.75}{$\pm$0.033}  & 31.70\scalebox{0.75}{$\pm$6.72}& 0.971\scalebox{0.75}{$\pm$0.028}    & 31.31\scalebox{0.75}{$\pm$6.57}& 0.970\scalebox{0.75}{$\pm$0.031}    \\ \Xhline{2\arrayrulewidth}
    \end{tabular}
\end{table}

\paragraph{Score-related baseline and our implementation} We implemented a simple sampling method with only ALD without our constraint term, which was named Langevin.
The score network we used was an implementation of NCSN++~\cite{Song2020ScoreBasedGM} trained on $4\times$NVIDIA Titan RTX GPUs with a total batch size of 64 for 150000 iterations. The learning rate also followed a cosine annealing schedule, which started from $2.5\times 10^{-5}$ and ended at $2.5\times 10^{-7}$. A warm-up of 4000 steps was adopted, where the learning rate rose from $0$ to $2.5\times 10^{-5}$ linearly. Gradient clipping was also used in training. The NCSN++ was optimized by AdamW optimizer.
Exponential moving average (EMA) was applied to the parameters of NCSN++ with momentum $m = 0.999$, and the final parameters produced by EMA were used for evaluation. We used 500 noise scales varying from $\sigma_{\text{min}} = 0.01$ to $\sigma_{\text{max}} = 100$, following the studies in~\cite{Song2020ImprovedTF}. All the sampling hyperparameters were tuned on the training dataset for 250 steps of Bayesian optimization via \texttt{ax-platform}. The results of each method were generated with the optimal parameters. The Langevin approach and our method shared the same score model. All the methods were implemented in Google's JAX~\cite{jax2018github} and TensorFlow~\cite{Abadi2016TensorFlowLM}. 

\begin{figure}[ht]
    \centering
    \subfigure[Uniform sampling]{
        \begin{minipage}[t]{0.31\textwidth}
            \centering
            \includegraphics[width=\textwidth]{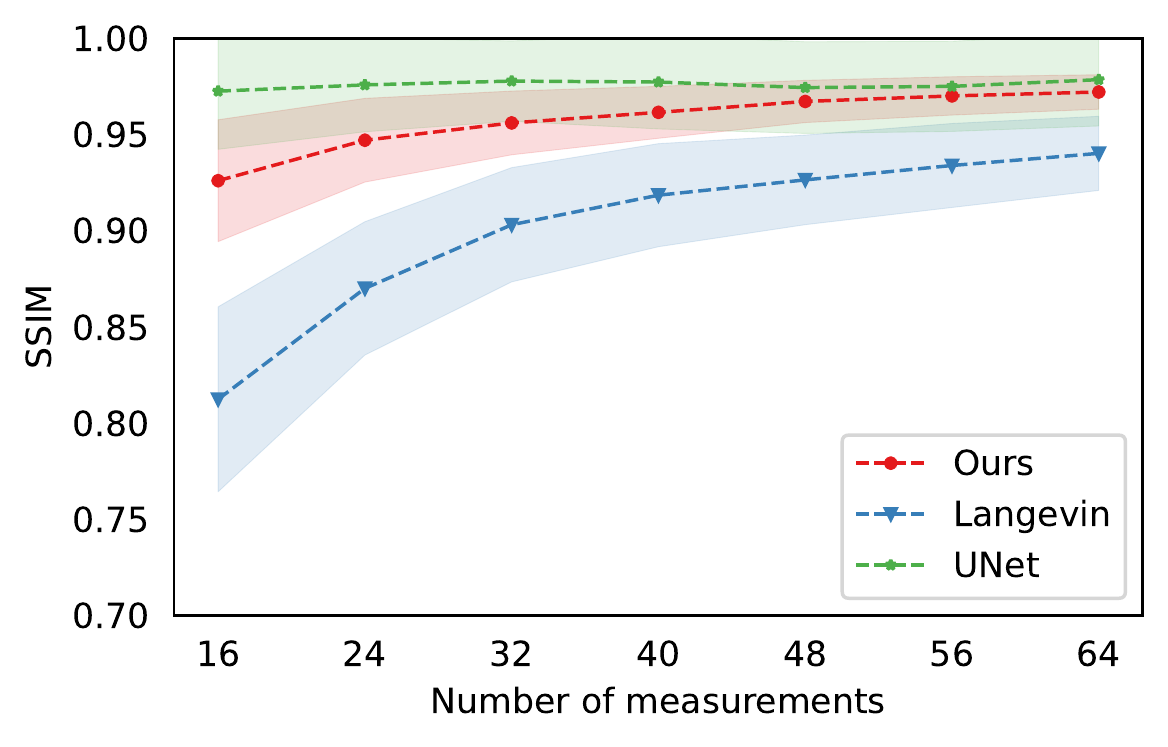} \\
            \includegraphics[width=\textwidth]{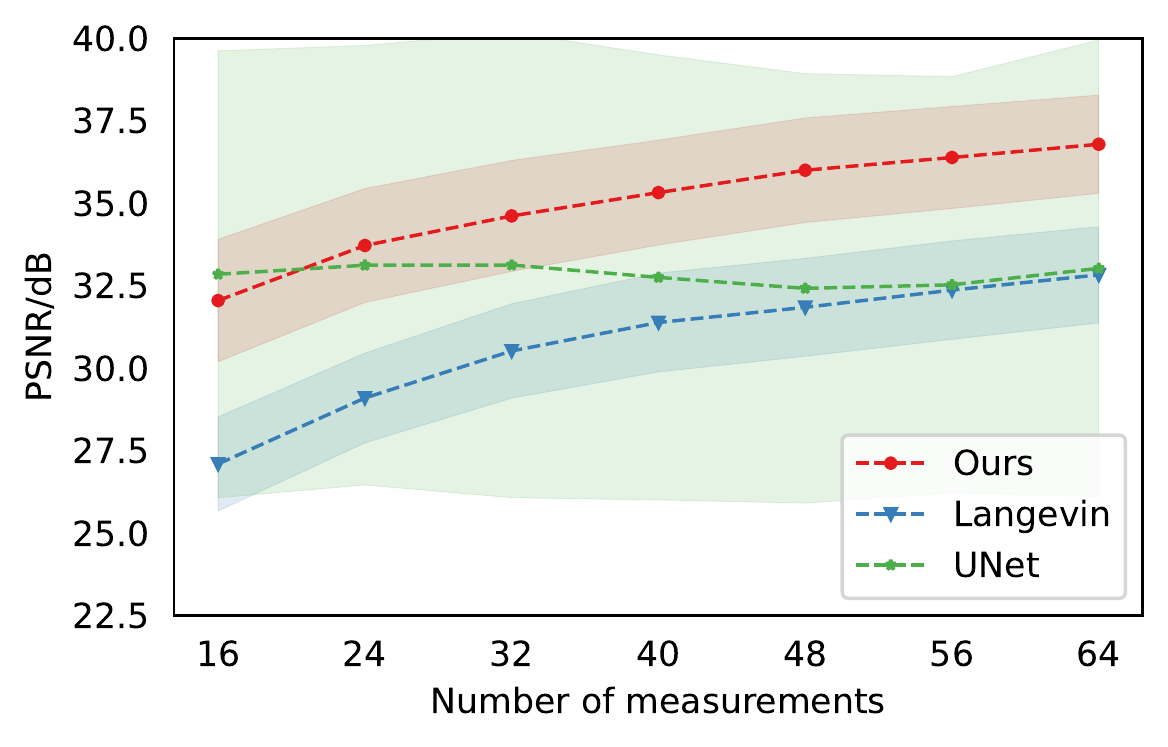}
        \end{minipage}
    }
    \subfigure[Random sampling]{
        \begin{minipage}[t]{0.31\textwidth}
            \centering
            \includegraphics[width=\textwidth]{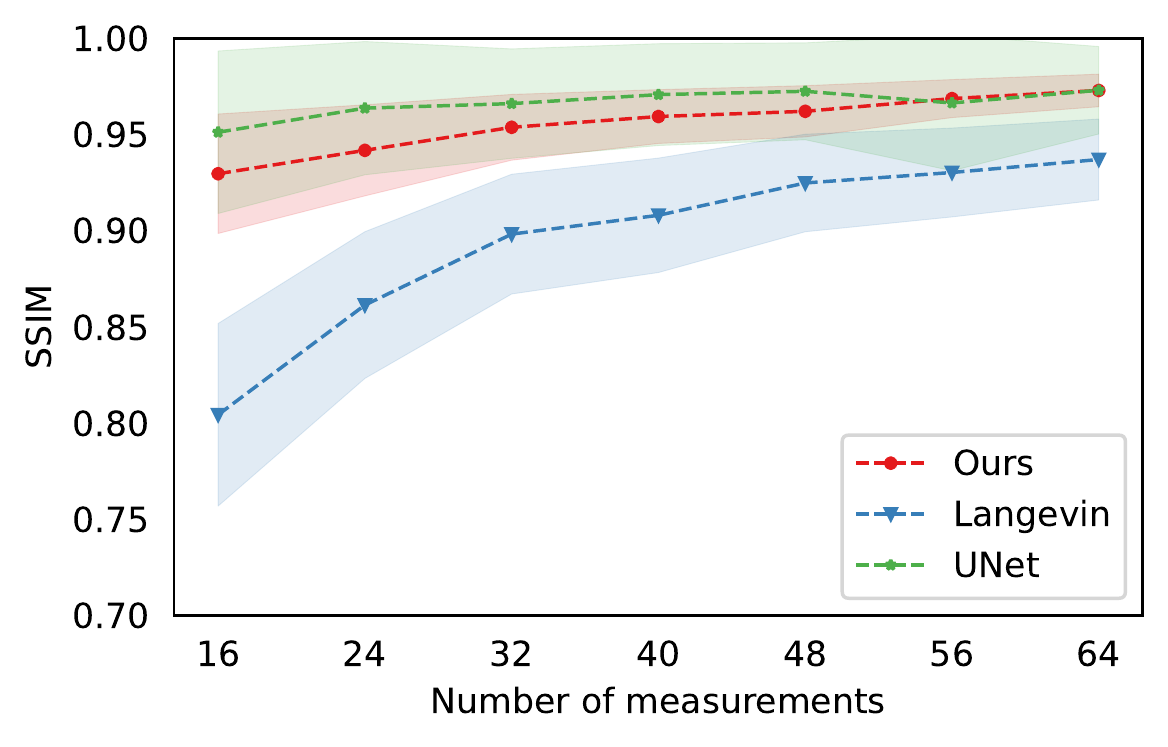} \\
            \includegraphics[width=\textwidth]{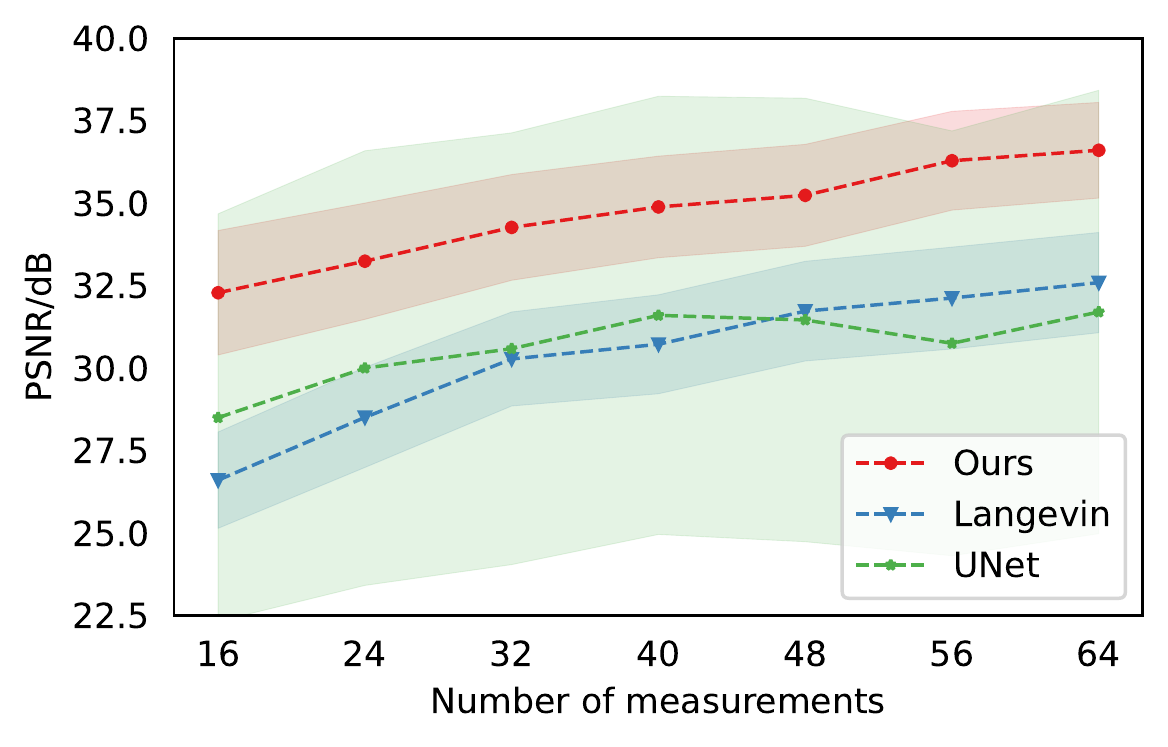}
        \end{minipage}
    }
    \subfigure[Limited view]{
        \begin{minipage}[t]{0.31\textwidth}
            \centering
            \includegraphics[width=\textwidth]{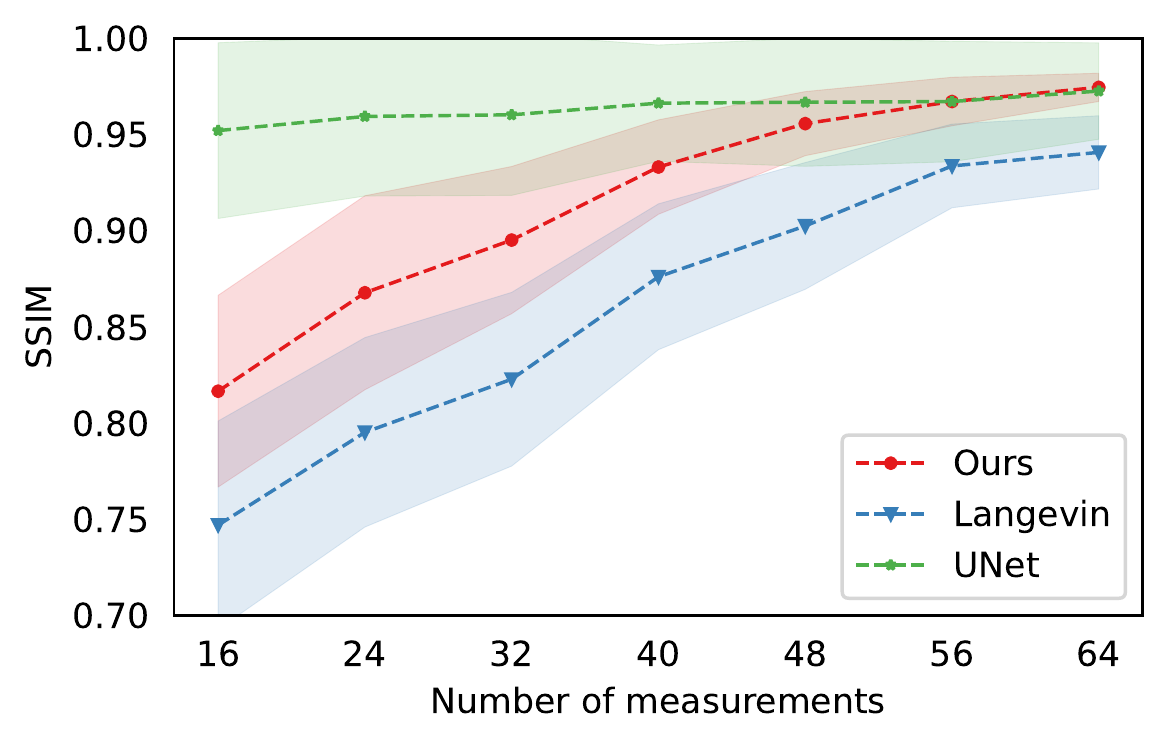} \\
            \includegraphics[width=\textwidth]{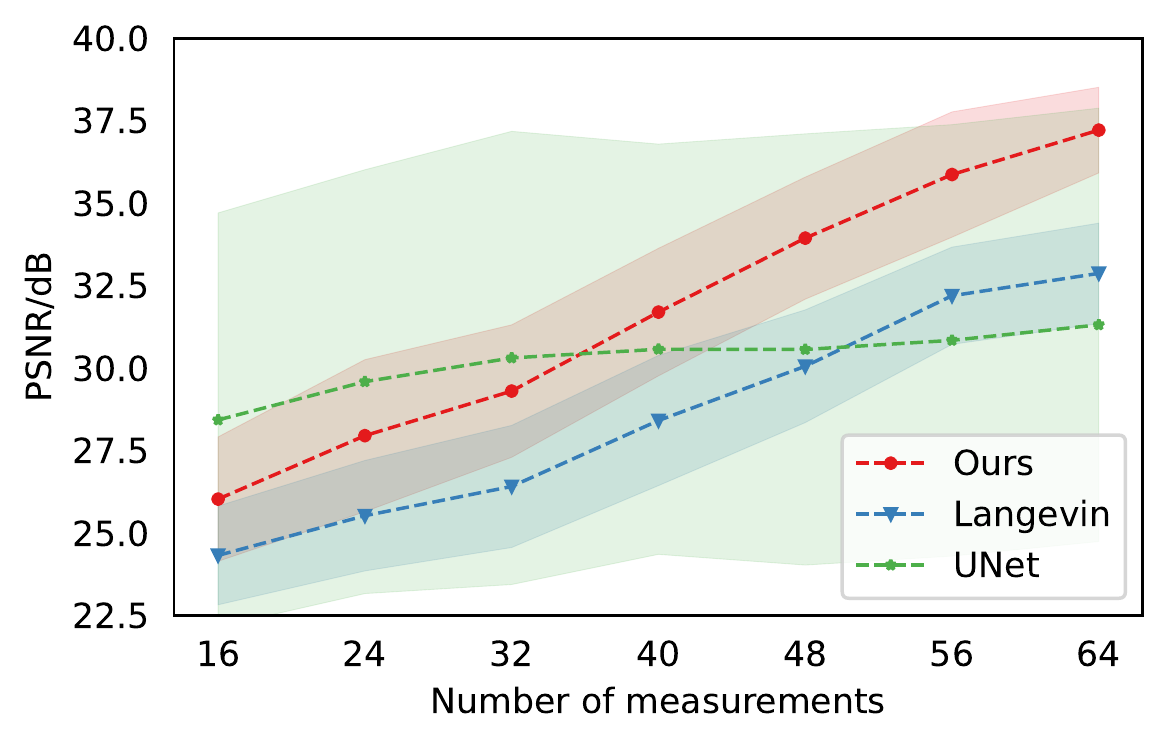}
        \end{minipage}
    }
    \caption{Results of the experiment reported in SSIM and PSNR}\label{fig:results-comparison}
\end{figure}
\section{Results and Discussions}\label{sec:limits}

Since our method is fully unsupervised, it is not restricted to any specific measurement process. 
We have tried different sampling patterns and different numbers of measurements, and all the results are shown in~\cref{fig:results-comparison} and~\cref{tab:results}. Note that the supervised counterparts were trained and evaluated on different sampling patterns and different numbers of measurements respectively.
As shown in~\cref{fig:results-comparison} and~\cref{tab:results}, we significantly outperformed the Langevin method in all measurement processes. In particular, the results of the Langevin approach showed significant distortion at 16 and 32 measurements. We achieved the highest PSNRs among all experiments with more than 40 measurements, while SSIMs still comparable. We could obtain competitive results to well-designed and trained supervised neural networks, while utilizing the same score model.

We have reported PSNR and SSIM values for they are basic metrics for estimating image reconstruction quality. 
Though the generalization capability is promising, the sampling time is relatively long due to high computation cost caused by excessive number of noise scales. The slow speed of reconstruction hinders the efficiency, which cannot meet the real-time scanning requirements of clinical needs. More in-depth studies are needed to modify the sampling algorithm for fast reconstruction.
\section{Conclusion}\label{sec:conclusion}

Collectively, we proposed \emph{the first} fully unsupervised framework named RCC-SGM for solving inverse problems in PAT via SGM and ALD with rotation consistency constraints. Evaluation results demonstrated that our method surpassed vanilla Langevin method in all kinds of measurement processes, even beat supervised counterparts under some particular circumstances. Besides, as a fully unsupervised method, we achieved higher generalization capability to new measurement processes than other approaches. If modifying the number of measurements and sampling patterns, we can use the same generative prior to generate high quality samples with a single score model.
\clearpage
\bibliographystyle{splncs04}
\bibliography{main}
\end{document}